\documentclass[letterpaper]{article} 
\usepackage{aaai2026}  
\usepackage{times}  
\usepackage{helvet}  
\usepackage{courier}  
\usepackage[hyphens]{url}  
\usepackage{graphicx} 
\usepackage{natbib}  
\usepackage{caption} 
\usepackage{algorithm}
\usepackage{algorithmic}

\usepackage{booktabs}  
\usepackage{amsmath}
\usepackage{amssymb}
\usepackage{multirow}

\usepackage{newfloat}
\usepackage{listings}
\DeclareCaptionStyle{ruled}{labelfont=normalfont,labelsep=colon,strut=off} 
\floatstyle{ruled}
\newfloat{listing}{tb}{lst}{}
\floatname{listing}{Listing}
\title{CausalCLIP: Causally-Informed Feature Disentanglement and Filtering for Generalizable Detection of Generated Images}
\author{
    Bo Liu\textsuperscript{\rm 1, 2},
    Qiao Qin\textsuperscript{\rm 1, 3},
    Qinghui He\textsuperscript{\rm 1, 3}\thanks{Corresponding author}
}
\affiliations{
    \textsuperscript{\rm 1} Chongqing Key Laboratory of Image Cognition, Chongqing University of Posts and Telecommunications, Chongqing, China \\
    \textsuperscript{\rm 2}School of Artificial Intelligence \\
    \textsuperscript{\rm 3}School of Computer Science and Technology

    boliu@cqupt.edu.cn, s240201048@stu.cqupt.edu.cn, d250201011@stu.cqupt.edu.cn
}

\begin{document}

\maketitle

\begin{abstract}
The rapid advancement of generative models has increased the demand for generated image detectors capable of generalizing across diverse and evolving generation techniques. However, existing methods, including those leveraging pre-trained vision-language models, often produce highly entangled representations, mixing task-relevant forensic cues (causal features) with spurious or irrelevant patterns (non-causal features), thus limiting generalization. To address this issue, we propose CausalCLIP, a framework that explicitly disentangles causal from non-causal features and employs targeted filtering guided by causal inference principles to retain only the most transferable and discriminative forensic cues. By modeling the generation process with a structural causal model and enforcing statistical independence through Gumbel-Softmax-based feature masking and Hilbert-Schmidt Independence Criterion (HSIC) constraints, CausalCLIP isolates stable causal features robust to distribution shifts. When tested on unseen generative models from different series, CausalCLIP demonstrates strong generalization ability, achieving improvements of 6.83\% in accuracy and 4.06\% in average precision over state-of-the-art methods. 
\end{abstract}


\section{Introduction}
The rapid development of generative models such as Generative Adversarial Networks (GANs)~\cite{gan1,gan2} and diffusion models~\cite{diffusion1} has drastically lowered the barrier to producing high-quality generated images. While these technologies hold great potential in applications such as image generation, editing, and enhancement, they also pose serious societal risks. Misuse of generative techniques can lead to the creation of hyper-realistic forged content, including manipulated faces, counterfeit evidence, and disinformation, which threatens public security, undermines media credibility, and challenges social governance. This growing threat has sparked an urgent need for a reliable and generalizable detector capable of identifying generated images across a wide range of generative models.
\begin{figure}[t]
    \centering
        \includegraphics[width=0.96\columnwidth]{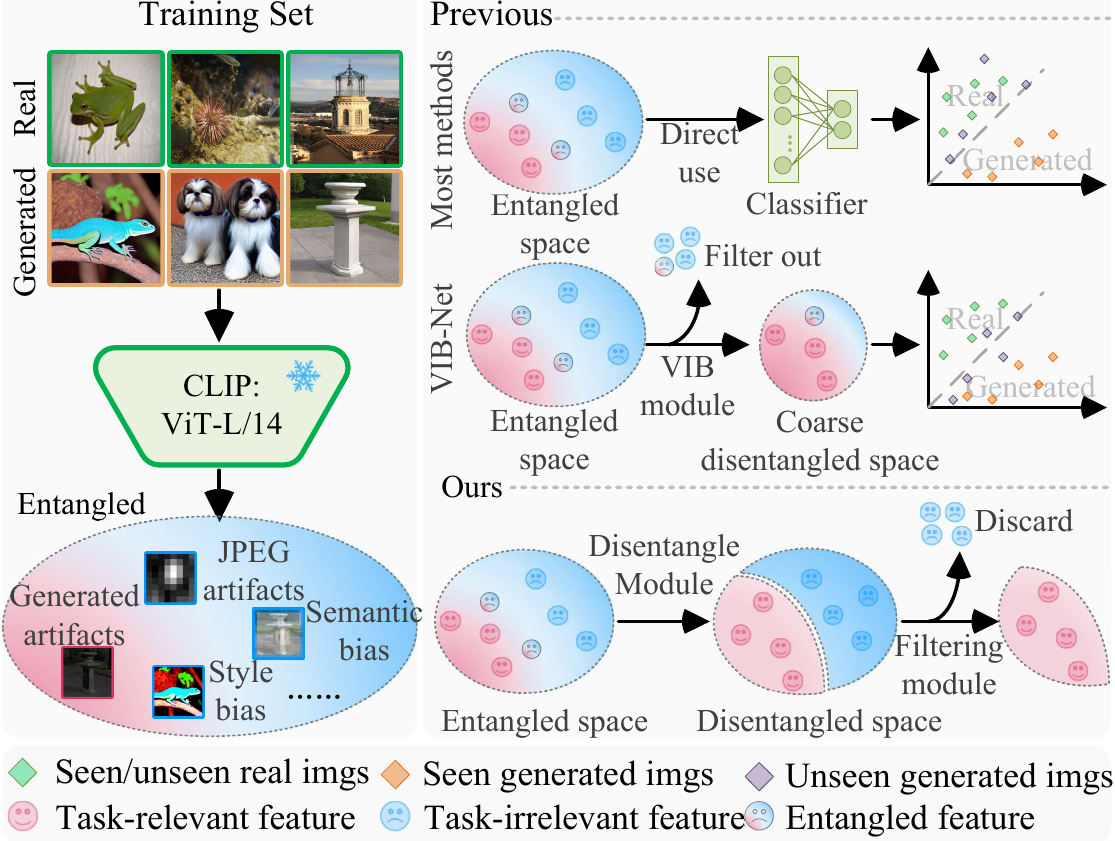}
        \caption{Comparison with previous methods for generated image detection. Prior approaches either directly use entangled CLIP features or filter them in the original space, leading to overfitting to spurious cues. Our method disentangles features into causal and non-causal features and performs filtering in the disentangled space, enabling better preservation of robust forensic cues and improved generalization to unseen generators.}
        \label{fig1}
\end{figure}

Early AI-generated image detection methods typically relied on supervised training of CNN-based classifiers~\cite{wang2020cnn,fredect}, where models learned generator-specific artifacts (e.g., upsampling traces or frequency anomalies). While these methods performed well on known generators, they exhibit severe performance degradation when facing unseen generation techniques due to overfiting to dataset-specific cues. To address this, recent research has explored leveraging pre-trained models, especially vision-language models like CLIP~\cite{clip_L_14}, which provide rich and semantically alignment representations with improved cross-domain transferability. Methods like CLIPping~\cite{clipping,raisingclip} adapt these features through linear probes, prompt tuning, or lightweight adapters to boost detection accuracy. However, even these methods still operate in highly entangled feature spaces, where causal forensic cues are mixed with spurious or non-causal patterns. VIB-Net~\cite{VIB} attempts to alleviate this issue by applying an information bottleneck to suppress task-irrelevant features, yet it does so without explicitly disentangling causal and non-causal features, leading to coarse filtering and suboptimal generalization under distribution shifts.

In this paper, we find that achieving strong cross-generator generalization requires explicitly separating causal forensic cues from spurious or training data-specific artifacts, rather than merely suppressing irrelevant features in an entangled representation space, which is a coarse filtering strategy that risks discarding task-relevant features, as shown in Fig.~\ref{fig1}. Building on this insight, we propose CausalCLIP, a causally guided framework that first disentangles causal features from non-causal features and then selectively leverages the causal subspace for detection. This disentangle-then-filter paradigm preserves stable and transferable forensic evidence that remains effective across diverse types of generative models, providing a theoretically grounded solution to overcoming the limitations of prior methods. Experimental results show that the proposed method achieves strong detection performance with remarkable cross-model generalization, demonstrating its effectiveness across a wide range of generative models.

We summarize our key contributions as follows:
\begin{itemize}
    \item We propose CausalCLIP, a detection framework following a disentangle-then-filter paradigm, which separates task-relevant features from task-irrelevant features to achieve stronger cross-model generalization.
    
    \item The proposed framework leverages adversarial disentanglement and counterfactual interventions to suppress non-causal features and preserve stable forensic cues for robust detection.

    \item When tested on unseen generative models from different series, CausalCLIP demonstrates strong generalization ability, achieving improvements of 6.83\% in accuracy and 4.06\% in average precision over state-of-the-art methods.
\end{itemize}

\section{Related Works}
\subsection{Generated Image Detection}
Although GANs (e.g., BigGAN~\cite{biggan}, ProGAN~\cite{progan}, and StyleGAN2/3~\cite{stylegan2,stylegan3}) have achieved remarkable image quality, they still leave subtle forensic cues, such as checkerboard patterns caused by upsampling and color inconsistencies. In contrast, diffusion models (e.g., ADM~\cite{ADM}, Glide~\cite{glide}, and Stable Diffusion~\cite{sdv14}) generate highly photorealistic images through iterative denoising, producing much fewer obvious artifacts but exhibiting distinct traces, such as over-smoothed textures and exaggerated style. The diversity and subtlety of these artifacts pose significant challenges for developing detectors that can generalize across different generative models.

\subsubsection{Generator-specific Detection}
Early methods for detecting AI-generated images primarily relied on supervised CNN-based classifiers, which learn to exploit generator-specific artifacts such as upsampling traces~\cite{wang2020cnn,watch}, frequency anomalies~\cite{fredect,closer,Spatialphase,liu2022eccv,liuonly}, and color inconsistencies~\cite{cross-band,Discovering}. Some more recent methods~\cite{npr,Secret_Lies_in_Color} also follow this strategy, leveraging visual artifacts or generator-specific cues to distinguish real and generated images. While these methods achieve high accuracy on known generators, they often fail to generalize to unseen generative models due to their strong dependence on specific features tied to the training distribution.

\subsubsection{Universal Generated Image Detection}
To further improve cross-generator generalization, recent research has turned to features extracted from large-scale pre-trained models, particularly CLIP~\cite{clip_L_14}, as universal feature extractors. UnivFD~\cite{Univfd} demonstrates that CLIP embeddings combined with lightweight classifiers can surpass previous detectors across both GAN-based and diffusion-based datasets. Raising~\cite{raisingclip} further reveals that frozen CLIP features, even when paired with simple classifiers, exhibit strong zero- and few-shot generalization. Building on this paradigm, CLIPping~\cite{clipping} adapts CLIP via prompt tuning and adapter-based fine-tuning, while C2P-CLIP~\cite{c2p} injects category-level prompts to better align image-text representations, thereby improving cross-domain detection performance. Moreover, the Few-Shot Learner~\cite{fewshoticml} extends this line of research by learning specialized metric-based detectors using only a handful of samples from a new generator. Despite these advances, CLIP-based methods still operate on entangled feature space, where task-relevant features are mixed with task-irrelevant features. VIB-Net~\cite{VIB} attempts to alleviate this by introducing an information bottleneck to suppress irrelevant features, but its filtering remains coarse and lacks explicit feature disentanglement, resulting in limited cross-generator generalization.

\begin{figure*}[t]
  \centering
  \includegraphics[width=0.95\textwidth]{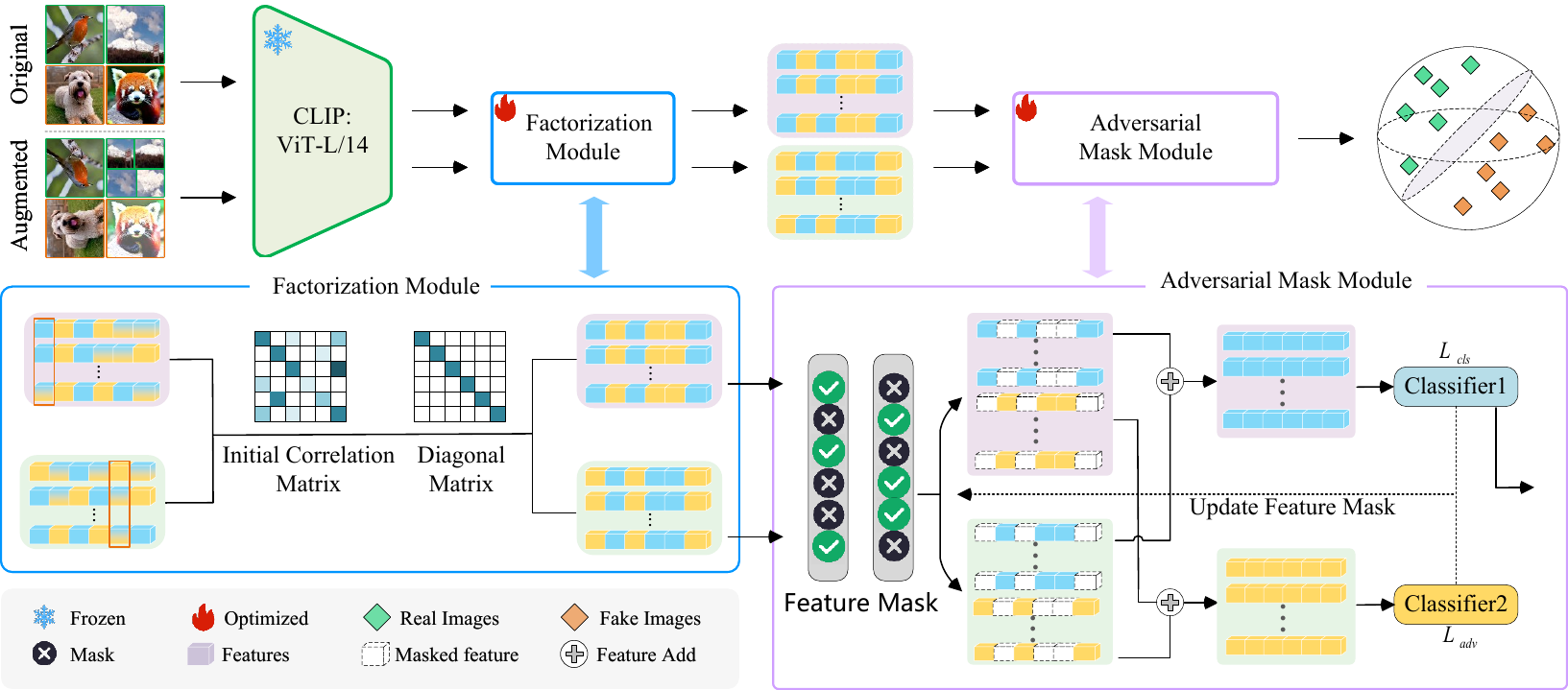}
  \caption{Architecture of the proposed CausalCLIP framework. An input image is processed by CLIP-ViT-L/14 to extract frozen features, which are disentangled by a Factorization Module into causal and non-causal components via correlation analysis. A Feature Mask is then learned by the Adversarial Mask Module to suppress non-causal features. Two binary classifiers are used: Classifier 1 predicts real vs. fake from masked features, while Classifier 2 acts adversarially, attempting to classify using the masked-out features. The mask is optimized to aid Classifier 1 while minimizing Classifier 2’s success, promoting the retention of stable, generation-invariant forensic features for better generalization.}
   \label{fig:architecture}
\end{figure*}

\subsection{ Representation Learning}
Causal representation learning (CRL) aims to improve generalization under distribution shifts by learning representations that align with underlying causal factors rather than spurious correlations~\cite{Toward}. In various domains, CRL has proven effective in disentangling task-relevant signals from irrelevant noise, thereby enhancing model interpretability and transferability~\cite{causality}. Typical approaches include adversarial training for disentanglement~\cite{disentangling}, structural causal models (SCMs) for explicitly modeling latent causal structures~\cite{causalvae}, and counterfactual interventions to reduce dataset bias and isolate stable features. For instance, CausalVAE~\cite{causalvae} leverages causal mechanisms to construct endogenous variables from exogenous noise, while DEAR~\cite{weakly} combines GANs with SCMs under weak supervision to relax independence assumptions and model causally entangled factors. These findings suggest that disentangling causal and non-causal features could help detection models focus on stable forensic evidence that remains effective across different generative models, addressing the common issue that features predictive on one generator often become unreliable when the generative mechanism changes.

\section{Method}
In this section, we introduce CausalCLIP, a causally-informed framework for generated image detection. We first provide an overview of the entire pipeline, as illustrated in Figure~\ref{fig:architecture}, and then describe each core module in detail. Specifically, we present the feature disentanglement (factorization) module, which separates causal and non-causal components of the extracted features, followed by the adversarial masking module, which enforces causal invariance through counterfactual interventions. Finally, we formulate the overall training objective and optimization strategy.

\subsection{Overview}
The key idea of CausalCLIP is to improve cross-model generalization in generated image detection by explicitly separating stable, task-relevant forensic cues (causal features) from non-causal features that are tied to specific datasets or generation styles. As illustrated in Figure~\ref{fig:architecture}, the input image is first processed by a frozen CLIP encoder to extract high-level semantic features. These features are then fed into the factorization module, which separates them into two complementary parts: causal features that capture intrinsic forensic cues stable across generation mechanisms, and non-causal features that encode generator- or dataset-specific artifacts. To further enhance generalization, the adversarial masking module performs targeted interventions on the non-causal features, weakening their influence and encouraging the classifier to focus on the causal features that remain reliable under distribution shifts. Finally, the refined causal features are passed to a lightweight classifier to determine whether the input image is real or generated.

\subsection{Factorization Module}
A major challenge in generated image detection is that CLIP embeddings entangle causal features, those truly indicative of real vs. fake, with non-causal artifacts specific to generators or datasets. This entanglement leads to overfitting and poor generalization to unseen generators. We address this with a Factorization Module that explicitly separates stable causal features from variable non-causal ones.

We assume that an image \(X \in \mathcal{X}\) can be explained by a structural causal model with two independent factors:
\begin{align}
Z_c &:= f_c(G, \epsilon_c), \\
Z_{nc} &:= f_{nc}(C, \epsilon_{nc}), \\
X &:= g(Z_c, Z_{nc}, \epsilon_x),
\end{align}
where \( Z_c \) represents causal features, while \( Z_{nc} \) captures non-causal features. Here, $G$ and $C$ are latent variables related to generation-independent content factors and generator-specific style or artifact factors, respectively, and $\epsilon_{\cdot}$ denotes exogenous noise.
 Our objective is to recover $Z_c$ from the entangled CLIP features while suppressing the influence of $Z_{nc}$.

Given the CLIP embedding \( E = \text{CLIP}(X) \in \mathbb{R}^d \), the factorization module learns a feature mask \( M \in [0,1]^d \) to separate:
\begin{equation}
\tilde{Z}_c = M \odot E, \quad \tilde{Z}_{nc} = (1 - M) \odot E,
\end{equation}
where \( \odot \) denotes element-wise multiplication. The mask \(M\) is parameterized by a Gumbel-softmax function:
\begin{equation}
M = \sigma((\text{MLP}(E) + g) / \tau), \quad g \sim \text{Gumbel}(0,1),
\end{equation}
with temperature \(\tau\) controlling the sparsity of feature selection. This mechanism ensures differentiable feature selection and yields a clean causal subspace $\tilde{Z}_c$ for downstream classification.

\subsection{Adversarial Masking Module}

Although the factorization module separates CLIP embeddings into causal and non-causal features, residual non-causal signals may still affect the classifier. To address this, we introduce an Adversarial Masking Module that uses an adversarial mechanism to ensure the decision boundary relies only on the stable causal subspace.

The adversarial design is motivated by the observation that explicitly suppressing non-causal features improves generalization under distribution shifts. We set up a minimax game where:
\begin{itemize}
    \item A classifier \( h \) predicts \( Y \in \{0,1\} \) (real vs.\ generated) based on the causal features \( \tilde{Z}_c \).
    \item An adversary \( d \) attempts to predict \( Y \) from the non-causal features \( \tilde{Z}_{nc} \).
\end{itemize}
The classifier and mask \( M \) are optimized to minimize classification loss while making \( \tilde{Z}_{nc} \) uninformative, forcing the model to rely solely on \( \tilde{Z}_c \).

The classification loss is defined as the standard binary cross-entropy:
\begin{equation}
\mathcal{L}_{cls}=-\mathbb{E}_{X,Y}[Y\log h(\tilde Z_c)+(1-Y)\log(1-h(\tilde Z_c))].
\end{equation}
where \( h(\tilde{Z}_c) \) denotes the predicted probability that \( X \) is generated.

The adversary is trained to maximize its ability to recover \( Y \) from non-causal features, while the mask and classifier are optimized to suppress such information. The adversarial loss is:
\begin{align}
\mathcal{L}_{adv}^{+} &= -\mathbb{E}_{X,Y}[ Y \log d(\tilde Z_{nc}) ], \\
\mathcal{L}_{adv}^{-} &= -\mathbb{E}_{X,Y}[ (1-Y) \log (1 - d(\tilde Z_{nc})) ].
\end{align}
\begin{equation}
\mathcal{L}_{adv} = \mathcal{L}_{adv}^{+} + \mathcal{L}_{adv}^{-}.
\end{equation}

To encourage a clear separation, we regularize the mask with sparsity and independence terms:
\begin{equation}
\mathcal{L}_{\text{mask}}
= \lambda_1 \|M\|_1
+ \lambda_2 \widehat{\text{HSIC}}(\tilde{Z}_c, \tilde{Z}_{nc}),
\end{equation}
where the \(\ell_1\)-norm promotes sparse feature selection, and the empirical HSIC term encourages statistical independence between the causal and non-causal subspaces. The HSIC is computed using Gaussian kernels with bandwidth selected via the median heuristic.

To enhance robustness, we introduce counterfactual interventions on causal features. By randomly masking a subset of dimensions,
\begin{equation}
\tilde{Z}_c^{CF} = \tilde{Z}_c \odot (1 - B),
\quad \text{where } B \sim \text{Bernoulli}(p),
\end{equation}
we simulate distributional perturbations and enforce prediction consistency via
\begin{equation}
\mathcal{L}_{\text{inv}}
= \text{KL}\big( h(\tilde{Z}_c) \parallel h(\tilde{Z}_c^{CF}) \big),
\end{equation}
compelling the classifier to rely on stable causal semantics rather than generator-dependent cues.

\subsection{Optimization Objective}

Our goal is to learn representations that preserve stable causal features while suppressing non-causal signals across generative models. We unify classification, adversarial, mask regularization, and counterfactual intervention losses as:
\begin{equation}
\mathcal{L}_{\text{total}} = \mathcal{L}_{\text{cls}} - \alpha \mathcal{L}_{\text{adv}} + \mathcal{L}_{\text{mask}} + \beta \mathcal{L}_{\text{inv}},
\end{equation}
where $\alpha$ and $\beta$ balance adversarial disentanglement and counterfactual consistency.

The objective is guided by two principles: (1) \textbf{Disentanglement}, where $\mathcal{L}_{\text{adv}}$ and $\mathcal{L}_{\text{mask}}$ separate causal from non-causal features; (2) \textbf{Stability}, where $\mathcal{L}_{\text{inv}}$ enforces consistent predictions under causal perturbations. Training alternates: $h$ minimizes $\mathcal{L}_{\text{cls}} + \beta \mathcal{L}_{\text{inv}}$, $d$ maximizes $\mathcal{L}_{\text{adv}}$, and $M$ minimizes $\mathcal{L}_{\text{total}}$, progressively refining causal features and suppressing non-causal ones.

\begin{table*}[t]
    \centering
    \renewcommand{\arraystretch}{1.2}
    \fontsize{9}{10.5}\selectfont
    \setlength{\tabcolsep}{1mm}  
\begin{tabular}{ccccccccccccccccc}
\hline
 & \multicolumn{7}{c}{Diffusion Models} & \multicolumn{6}{c}{Generative Adversarial Networks} & \multicolumn{2}{c}{Others} & \textit{AP} \\ 
   \cmidrule(r){2-8}
   \cmidrule(r){9-14} 
   \cmidrule(r){15-16}
   \cmidrule(r){17-17} 

\multirow{-2}{*}{Method} & SD1.4 & SD1.5 & ADM & GLIDE & Midj & Wukong & VQDM & \begin{tabular}[c]{@{}c@{}}Pro-\\ GAN\end{tabular} & \begin{tabular}[c]{@{}c@{}}Cycle-\\ GAN\end{tabular} & \begin{tabular}[c]{@{}c@{}}Big-\\ GAN\end{tabular} & \begin{tabular}[c]{@{}c@{}}Style-\\ GAN\end{tabular} & \begin{tabular}[c]{@{}c@{}}Star-\\ GAN\end{tabular} & \begin{tabular}[c]{@{}c@{}}Gau-\\ GAN\end{tabular} & \begin{tabular}[c]{@{}c@{}}Deep-\\ fake\end{tabular} & SAN & Avg \\ \hline

CNNSpot & \underline{99.98} & 99.83 & 51.10 & 58.80 & 67.93 & 99.80 & 49.92 & 53.15 & 50.23 & 50.22 & 49.79 & 47.07 & 56.08 & 54.86 & 54.03 & 63.24 \\
Fusing & 99.90 & 97.98 & 69.30 & 94.20 & 81.20 & 99.90 & 84.60 & 67.63 & 87.79 & 64.84 & 69.37 & \underline{91.20} & 43.08 & 72.56 & 89.42 & 81.07 \\
Lgrad & 99.94 & 99.92 & 58.52 & 84.00 & 91.06 & 99.72 & 56.34 & 83.59 & 90.24 & 56.49 & 47.51 & 69.93 & 49.25 & 66.49 & 65.09 & 78.24 \\
Univfd & 96.04 & 96.26 & 66.34 & 93.73 & 91.06 & 90.98 & 74.53 & 51.77 & 63.42 & 72.00 & 75.81 & 54.12 & 65.99 & 70.24 & 83.34 & 75.31 \\
NPR & \textbf{100.00} & \underline{99.97} & 94.70 & 95.80 & 95.50 & \textbf{100.00} & \underline{86.30} & 83.30 & 94.90 & \underline{89.60} & 72.00 & 82.70 & 66.00 & \textbf{85.30} & \underline{95.90} & 89.98 \\
CLIPping & 93.97 & 93.10 & 68.00 & 87.44 & 77.34 & 86.52 & 77.17 &  88.54 & 88.44 & 61.64 & 85.33 & 77.23 & 81.56 & 61.19 & 57.37 & 80.87 \\
VIB-Net & \textbf{100.00} & \underline{99.97} & \textbf{95.49} & \underline{97.13} & \underline{97.81} & 99.93 & 83.02 & \underline{96.59} & \textbf{98.44} & 81.50 & \textbf{97.17} & 84.31 & \textbf{96.94} & \underline{81.32} & 93.27 & \underline{94.60} \\
Ours & \textbf{100.00} & \textbf{99.99} & \underline{95.37} & \textbf{99.47} & \textbf{98.23} & \underline{99.95} & \textbf{98.82} & \textbf{97.33} & \underline{98.35} & \textbf{97.43} & \underline{96.29} & \textbf{98.62} & \underline{96.84} & 79.35 & \textbf{97.79} & \textbf{96.92} \\
\hline
\end{tabular}
\caption{ The \textit{AP} values of different methods trained on Diffusion source images. Data in bold represents the best, while data underlined represents the second best.}
\label{tab:diffusion_ap}
\end{table*}

\begin{table*}[t]
    \centering
    \renewcommand{\arraystretch}{1.2}
    \setlength{\tabcolsep}{1mm}
    \fontsize{9}{10.5}\selectfont
\begin{tabular}{ccccccccccccccccc}
\hline
 & \multicolumn{7}{c}{Diffusion Models} & \multicolumn{6}{c}{Generative Adversarial Networks} & \multicolumn{2}{c}{Others} & \textit{ACC} \\ 
   \cmidrule(r){2-8}
   \cmidrule(r){9-14} 
   \cmidrule(r){15-16}
   \cmidrule(r){17-17} 

\multirow{-2}{*}{Method} & SD1.4 & SD1.5 & ADM & GLIDE & Midj & Wukong & VQDM & \begin{tabular}[c]{@{}c@{}}Pro-\\ GAN\end{tabular} & \begin{tabular}[c]{@{}c@{}}Cycle-\\ GAN\end{tabular} & \begin{tabular}[c]{@{}c@{}}Big-\\ GAN\end{tabular} & \begin{tabular}[c]{@{}c@{}}Style-\\ GAN\end{tabular} & \begin{tabular}[c]{@{}c@{}}Star-\\ GAN\end{tabular} & \begin{tabular}[c]{@{}c@{}}Gau-\\ GAN\end{tabular} & \begin{tabular}[c]{@{}c@{}}Deep-\\ fake\end{tabular} & SAN & Avg \\ \hline

CNNSpot & 99.48 & 93.35 & 50.10 & 50.90 & 56.42 & 97.90 & 50.04 & 50.27 & 49.81 & 50.10 & 50.98 & 49.77 & 50.38 & 51.98 & 50.22 & 60.51 \\
Fusing & \underline{99.90} & \textbf{99.91} & 51.30 & 57.50 & 52.30 & \underline{99.90} & 64.20 & 51.20 & 52.40 & 53.50 & 50.20 & 58.20 & 49.32 & 51.02 & 64.48 & 63.71 \\
Lgrad & 99.12 & 99.05 & 53.00 & 64.24 & 76.34 & 97.53 & 50.93 & 61.61 & 60.74 & 48.82 & 61.43 & 50.17 & 49.70 & 50.17 & 56.49 & 65.29 \\
Univfd & 83.55 & 84.80 & 53.35 & 75.30 & 71.60 & 73.55 & 55.10 & 58.65 & 59.30 & 61.45 & 56.80 & 61.45 & 55.30 & 58.40 & 72.00 & 65.37 \\
NPR & \textbf{100.00} & \underline{99.90} & 73.00 & \textbf{89.70} & \underline{82.30} & \textbf{100.00} & 68.30 & 60.30 & 67.20 & 59.20 & 58.00 & 73.20 & 52.00 & \textbf{74.80} & \textbf{89.60} & 76.50 \\
CLIPping & 96.07 & 95.48 & 70.14 & 85.00 & 77.66 & 88.86 & 79.35 & 88.78 & 88.48 & 89.57 & \underline{80.69} & \underline{90.82} & 85.94 & 66.91 & 61.64 & 83.03 \\
VIB-Net & 99.55 & 99.20 & \underline{73.85} & 74.25 & \textbf{88.05} & 98.25 & \underline{89.35} & \underline{89.70} & \underline{88.60} & \underline{91.20} & 74.10 & 80.70 & \underline{87.15} & \underline{72.00} & 81.50 & \underline{85.83} \\
Ours & 99.81 & 99.59 & \textbf{77.95} & \underline{89.22} & 80.72 & 99.08 & \textbf{90.73} & \textbf{91.22} & \textbf{97.35} & \textbf{94.27} & \textbf{87.78} & \textbf{95.75} & \textbf{94.36} & 71.64 & \underline{87.21} & \textbf{90.45} \\
\hline
\end{tabular}
\caption{ The \textit{ACC} values of different methods trained on Diffusion source images. Similar to the symbols in Table \ref{tab:diffusion_ap}.}
\label{tab:diffusion_acc}
\end{table*} 

\section{Experiments}
In this section, we evaluate the proposed CausalCLIP framework on various datasets and compare it with SOTA generated image detection methods. We first introduce the training and testing datasets and evaluation metrics, followed by implementation details.

\subsection{Experimental Setting}
\subsubsection{Training Datasets}
To evaluate the detection performance and generalization performance of the proposed method across different generative models. We use two representative datasets for training. The first is ProGAN from CNNDet~\cite{wang2020cnn}, which contains 360k real images from LSUN~\cite{lsun} and 360k generated images from ProGAN. The second is Stable Diffusion v1.4 from GenImage~\cite{genimage}, which includes 32k high-quality generated images from Stable Diffusion v1.4~\cite{sdv14}, paired with their corresponding real images. All methods are trained on either ProGAN or SDv1.4 datasets to evaluate cross-generator generalization.

\subsubsection{Testing Datasets}
We consider 15 testing datasets covering both GAN-based and diffusion-based models. The ForenSynths test set includes images generated by six different GAN-based models, representing a variety of architectures such as ProGAN~\cite{gan1}, CycleGAN~\cite{cyclegandataset}, StarGAN~\cite{stargan}, StyleGAN~\cite{cyclegandataset}, BigGAN~\cite{biggan}, and GauGAN~\cite{gaugan}. In addition, two other models, DeepFake~\cite{deepfake} and SAN~\cite{san}, are also included. The corresponding real images are sampled from six commonly used datasets, namely ImageNet~\cite{imagenet}, LSUN~\cite{lsun}, FaceForensics++~\cite{faceforensics++}, CelebA-HQ~\cite{gan2}, CelebA~\cite{cele}, and COCO~\cite{coco}. The GenImage test set contains images generated by seven state-of-the-art diffusion models, including Stable Diffusion v1.4, v1.5~\cite{sdv14}, VQDM~\cite{vqdm}, ADM~\cite{ADM}, GLIDE~\cite{glide}, Wukong, and Midjourney. Each generated image is paired with a corresponding real image to ensure fair and comparable evaluation.

\subsubsection{Evaluation Metrics}
We adopt two commonly used metrics: Average Precision (\textit{AP}) and Accuracy (\textit{ACC}). \textit{AP} is a threshold-independent metric that provides a comprehensive evaluation of the model’s performance across different decision thresholds. In contrast, \textit{ACC} measures the proportion of correctly classified samples among all test samples and is sensitive to the choice of classification threshold.

\subsubsection{Baselines}
We compare the proposed CausalCLIP with a diverse set of SOTA detection methods, including {CNNDetection} (CVPR 2020)\cite{wang2020cnn}, {FreDect} (ICML 2020)\cite{fredect}, {Fusing} (ICIP 2022)\cite{2022fusing}, {LGrad} (CVPR 2023)\cite{lgrad}, {DIRE} (CVPR 2023)\cite{dire}, {UnivFD} (CVPR 2023)\cite{Univfd}, {NPR} (CVPR 2024)\cite{npr}, {CLIPping} (ICMR 2024)\cite{clipping}, and {VIB-Net} (CVPR 2025)~\cite{VIB}.

\subsubsection{Implementation Details}
We employ the frozen image encoder of the pretrained CLIP-ViT/L-14 model as the backbone to extract high-level semantic features. On top of the extracted features, we train the downstream causal representation module and the discriminative classifier. We use the Adam optimizer with an initial learning rate of $1\times10^{-4}$ and a batch size of 256, with early stopping applied to prevent overfitting. All input images are center-cropped to $224\times224$ pixels. The entire framework is implemented in PyTorch, and all experiments are conducted on a NVIDIA Tesla V100 GPU.

\begin{table*}[t]
\centering
\renewcommand{\arraystretch}{1.2}
\fontsize{9}{10.5}\selectfont
\setlength{\tabcolsep}{1mm}
\begin{tabular}{cccccccccccccccccc}
\hline
 & \multicolumn{6}{c}{Generative Adversarial Networks} & \multicolumn{2}{c}{Others} & \multicolumn{7}{c}{Diffusion Models} & \textit{AP} \\ 
 \cmidrule(r){2-7}
 \cmidrule(r){8-9}
 \cmidrule(r){10-16}
 \cmidrule(r){17-17}
\multirow{-2}{*}{Methods} 
& \begin{tabular}[c]{@{}c@{}}Pro-\\GAN\end{tabular} 
& \begin{tabular}[c]{@{}c@{}}Cycle-\\GAN\end{tabular}
& \begin{tabular}[c]{@{}c@{}}Big-\\GAN\end{tabular}
& \begin{tabular}[c]{@{}c@{}}Stytle-\\GAN\end{tabular}
& \begin{tabular}[c]{@{}c@{}}Star-\\GAN\end{tabular}
& \begin{tabular}[c]{@{}c@{}}Gau-\\GAN\end{tabular}

& \begin{tabular}[c]{@{}c@{}}Deep-\\fake\end{tabular} & SAN 
& SD1.4 & SD1.5 & ADM & GLIDE & Midj & Wukong & VQDM & Avg \\ \hline

CNNSpot & \underline{99.99} & 96.40 & 87.50 & 96.94 & 94.24 & 98.28  & 64.42 & 55.89 & 52.86 & 53.25 & 65.14 & 68.10 & 56.60 & 51.15 & 69.49 & 74.02 \\
FreDect & \underline{99.99} & 84.77 & 93.62 & 88.97 & 99.48 & 82.85  & 70.77 & 49.50 & 38.50 & 38.41 & 63.72 & 54.73 & 47.25 & 40.44 & 86.01 & 69.27 \\
Fusing & \textbf{100.00} & 95.50 & 90.76 & 99.48 & 99.82 & 88.32 & 71.12 & 77.33 & 65.30 & 65.62 & 74.85 & 77.44 & 69.91 & 64.53 & 75.42 & 81.03 \\
Lgrad & 99.90 & 94.01 & 90.75 & \textbf{99.80} & \textbf{99.98} & 79.29  & 71.71 & 45.09 & 70.90 & 71.72 & 71.83 & 75.96 & 71.42 & 66.51 & 70.23 & 78.61 \\
Univfd & \textbf{100.00} & 99.21 & 98.31 & 97.98 & 99.35 & \underline{99.80}  & 82.04 & 82.18 & 85.48 & 82.30 & 84.34 & 84.04 & 69.10 & 90.13 & 94.96 & 89.85 \\
NPR & \textbf{100.00} & 98.50 & 87.80 & \underline{99.80} & \underline{99.90} & 85.50  & 82.40 & 71.60 & 84.00 & \underline{84.60} & 74.60 & 85.70 & \textbf{85.40} & 80.50 & 81.20 & 86.77  \\
CLIPping & 99.85 & 94.03 & 91.27 & 95.24 & 99.05 & 89.52  & 73.89 & 56.62 & 58.28 & 57.85 & 77.25 & 77.93 & 52.43 & 60.65 & 80.00 & 77.59  \\
VIB-Net & \textbf{100.00} & \textbf{99.80} & \underline{99.29} & 98.79 & 99.72 & \textbf{99.99}  & \textbf{92.64} & \underline{91.62} & \underline{87.24} & \textbf{86.98} & \underline{87.88} & \underline{88.53} & 75.68 & \underline{90.92} & \underline{96.51} & \underline{93.04} \\
Ours & \textbf{100.00} & \underline{99.27} & \textbf{99.31} & 99.70 & 99.89 & 99.72 & \underline{90.78} & \textbf{93.14} & \textbf{89.91} & 87.59 & \textbf{88.92} & \textbf{92.41} & \underline{82.72} & \textbf{93.52} & \textbf{97.12} & \textbf{94.27} \\
\hline
\end{tabular}
\caption{The \textit{AP} values of different methods trained on GAN source images. Similar to the symbols in Table \ref{tab:diffusion_ap}.}
\label{tab:progan_ap}
\end{table*}

\begin{table*}[t]
\centering
\renewcommand{\arraystretch}{1.2}
\fontsize{9}{10.5}\selectfont
\setlength{\tabcolsep}{1mm}
\begin{tabular}{cccccccccccccccccc}
\hline
 & \multicolumn{6}{c}{Generative Adversarial Networks} & \multicolumn{2}{c}{Others} & \multicolumn{7}{c}{Diffusion Models} & \textit{ACC} \\ 
 \cmidrule(r){2-7}
 \cmidrule(r){8-9}
 \cmidrule(r){10-16}
 \cmidrule(r){17-17}
\multirow{-2}{*}{Methods} 
& \begin{tabular}[c]{@{}c@{}}Pro-\\GAN\end{tabular} 
& \begin{tabular}[c]{@{}c@{}}Cycle-\\GAN\end{tabular}
& \begin{tabular}[c]{@{}c@{}}Big-\\GAN\end{tabular}
& \begin{tabular}[c]{@{}c@{}}Style-\\GAN\end{tabular}
& \begin{tabular}[c]{@{}c@{}}Star-\\GAN\end{tabular}
& \begin{tabular}[c]{@{}c@{}}Gau-\\GAN\end{tabular}
& \begin{tabular}[c]{@{}c@{}}Deep-\\fake\end{tabular} & SAN 
& SD1.4 & SD1.5 & ADM & GLIDE & Midj & Wukong& VQDM & Avg \\ \hline

CNNSpot & \underline{99.99} & 87.59 & 71.18 & 89.95 & 94.60 & 81.44 & 51.69 & 50.00 & 50.82 & 50.88 & 60.20 & 57.85 & 50.77 & 51.13 & 56.20 & 66.95  \\
FreDect & 99.36 & 78.76 & 81.97 & 78.01 & 94.62 & 80.57 & 63.29 & 50.00 & 40.02 & 40.38 & 64.66 & 55.43 & 46.89 & 41.54 & 78.95 & 66.30  \\
Fusing & 99.90 & 87.01 & 77.32 & 85.21 & 97.04 & 76.95 & 53.76 & 54.56 & 51.04 & 51.34 & 56.50 & 57.15 & 52.12 & 51.67 & 55.09 & 67.11  \\
Lgrad & 99.80 & 86.94 & 85.63 & 91.08 & 99.27 & 72.49 & 56.42 & 44.47 & 63.03 & 63.67 & 67.10 & 66.10 & 56.20 & 63.60 & 67.02 & 72.19  \\
Univfd & 99.90 & 98.50 & 94.50 & 84.40 & 95.85 & 99.50 & 67.40 & 56.50 & 63.10 & 63.57 & 66.90 & 61.70 & 57.85 & 71.06 & 85.00 & 77.72  \\
NPR & 99.80 & 96.10 & 84.40 & \textbf{97.70} & \underline{99.30} & 82.50 & \underline{80.20} & 69.20 & 76.60 & \underline{77.90} & 69.70 & 77.30 & \textbf{77.80} & \underline{76.10} & 78.10 & 82.85  \\
CLIPping & 99.88 & 96.74 & \underline{94.77} & \underline{94.87} & 99.47 & \underline{94.15} & 76.48 & 60.27 & 61.26 & 60.78 & \textbf{80.77} & \underline{80.69} & 53.95 & 63.94 & 84.76 & 80.19  \\
VIB-Net & \underline{99.99} & \underline{99.00} & \textbf{95.75} & 91.25 & 98.95 & \textbf{99.70} & \textbf{83.20} & \underline{70.50} & \underline{71.55} & 70.00 & 71.45 & 69.40 & 61.25 & 75.90 & \underline{86.65} & \underline{82.97} \\
Ours & 99.06 & \textbf{99.95} & 91.73 & 92.73 & 96.14 & 95.92 & 72.34 & \textbf{79.45} & \textbf{82.58} & \textbf{82.54} & \underline{78.94} & \textbf{82.73} & \underline{65.99} & \textbf{85.91} & \textbf{87.48} & \textbf{86.23} \\
\hline
\end{tabular}
\caption{The \textit{ACC} values of different methods trained on GAN source images. Similar to the symbols in Table \ref{tab:diffusion_ap}.}
\label{tab:progan_acc}
\end{table*}

\subsection{Cross-Model Generalization Evaluation}
To comprehensively assess the generalization capability of CausalCLIP under distribution shifts, we design two challenging cross-model detection experiments. Specifically, we train the model on images generated by either a diffusion model or a GAN, and test it on a diverse set of generative models. This setup mimics real-world scenarios where training and testing distributions differ significantly due to variations in generation mechanisms and visual styles.

\subsubsection{Diffusion-Sources Evaluation}
In the first setting, we train CausalCLIP on images generated by Stable Diffusion v1.4 and evaluate its performance on samples from 15 other generative models. As shown in Tables~\ref{tab:diffusion_ap} and~\ref{tab:diffusion_acc}, many existing detection methods (e.g., LGrad, UnivFD) suffer substantial performance drops on these newer diffusion models, with declines exceeding 40\% in both \textit{AP} and \textit{ACC}. This indicates that these methods often overfit to non-causal artifacts in the training data, such as stylistic inconsistencies or generator-specific textures, which fail to generalize under distributional shifts. 

In contrast, CausalCLIP consistently maintains strong performance across all models, achieving improvements of 2.32\% in \textit{AP} and 4.62\% in \textit{ACC}. When further tested on unseen GAN-based generators, it achieves additional improvements of 6.83\% in \textit{ACC} and 4.06\% in \textit{AP}, confirming strong cross-family generalization. These gains are attributed to CausalCLIP’s causality-guided representation learning, where the causal mask and adversarial intervention collaboratively disentangle causal features and foster counterfactually invariant representations for robust generalization.

\subsubsection{GAN-Sources Evaluation}
To further assess the generalization ability of CausalCLIP under pronounced style variations, we conduct a second experiment where the model is trained on images generated by ProGAN and evaluated on 15 diverse generative models.     Due to the substantial differences in generative characteristics between ProGAN and modern diffusion models, this setting introduces a more severe distribution gap, making it well-suited for examining the model’s robustness to both style and semantic shifts.

Results in Tables~\ref{tab:progan_ap} and~\ref{tab:progan_acc} demonstrate that many baseline methods (e.g., Fusing, LGrad, UnivFD) suffer severe performance drops on the new diffusion models, with \textit{AP} and \textit{ACC} often falling below 60\% and, in some cases, plummeting to 55\% on models such as ADM, Midjourney, and VQDM. These results highlight the heavy reliance of conventional methods on generator-specific artifacts, which fail to transfer across different generation paradigms. In contrast, CausalCLIP consistently outperforms all baselines, achieving improvements of 1.23\% in \textit{AP} and 3.26\% in \textit{ACC}. When further tested on unseen Diffusion-based generators, it achieves additional improvements of 8.57\% in \textit{ACC} and 2.64\% in \textit{AP}, confirming strong generalization.  

\subsection{Feature Disentanglement}

\subsubsection{Feature Space Visualization} 
\begin{figure}[t]
\centering
\includegraphics[width=0.96\columnwidth]{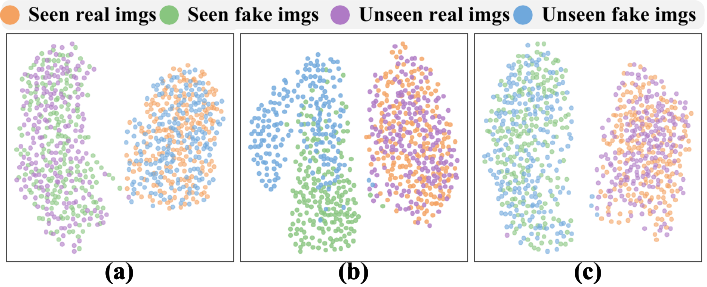}
\caption{
    UMAP visualization of real and fake image features under seen and unseen settings. (a) CLIP shows strong domain entanglement, (b) VIB achieves partial separation, (c) our method provides clear separation across all domains.
    }
 \label{fig:visualization_compare}
\end{figure}

To assess the discriminative power of different representations, we visualize features using UMAP for both seen and unseen generators. As shown in Figure~\ref{fig:visualization_compare}, CLIP features exhibit strong entanglement across domains, making real–fake discrimination difficult. VIB achieves partial separation but still overlaps in unseen cases. In contrast, our method achieves clear separation across all domains, indicating superior disentanglement and generalization.

\subsubsection{Robustness} 
\begin{figure}[t]
    \centering
    \includegraphics[width=0.96\columnwidth]{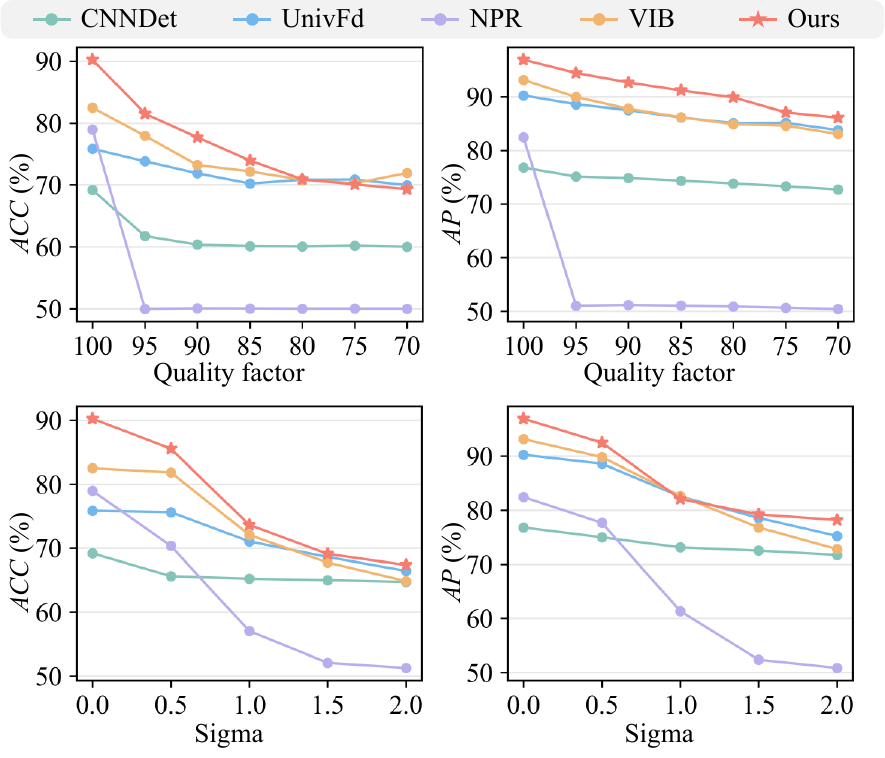}
    \caption{
Robustness analysis under JPEG compression (top) and Gaussian blur (bottom). Our method (star markers) shows better stability under most perturbations.
}
    \label{fig:robustness}
\end{figure}

We further evaluate model robustness against two common input perturbations: JPEG compression and Gaussian blur. Figure~\ref{fig:robustness} reports accuracy and average precision under varying quality factors and blur levels. Conventional approaches degrade significantly as perturbations increase, while our method maintains the most stable performance across all settings.

\begin{table}[t]
    \centering
    \renewcommand{\arraystretch}{1.2}
    \setlength{\tabcolsep}{1mm}
    \begin{tabular}{cc cc  cc}
    \toprule
    \multirow{2}{*}{\begin{tabular}[c]{@{}c@{}}Factorization\\module\end{tabular}} &
    \multirow{2}{*}{\begin{tabular}[c]{@{}c@{}}Masking\\module\end{tabular}} &
    \multicolumn{2}{c}{\textit{ACC}} &
    \multicolumn{2}{c}{\textit{AP}} \\ 
     \cmidrule(r){3-4}  \cmidrule(r){5-6}
     &  & OP & GP & OP & GP \\
    \midrule
    $\times$      & $\times$      & 65.37 & 60.42 & 75.31 & 67.09 \\
    $\checkmark$  & $\times$      & 79.42 & 75.91 & 89.53 & 87.78 \\ 
    $\times$      & $\checkmark$  & 70.73 & 65.94 & 82.13 & 79.28 \\
    $\checkmark$  & $\checkmark$  & 90.45 & 89.95 & 96.92 & 95.25 \\
    \bottomrule
    \end{tabular}
    \caption{Ablation study on the contributions of the disentanglement and masking modules. ``$\times$'' indicates the module is disabled, while ``$\checkmark$'' indicates it is enabled. The baseline UnivFD corresponds to the case where both modules are disabled. Combining both modules achieves the best results.}
    \label{tab:module_ablation}
\end{table}

\subsection{Ablation Studies}
To evaluate the individual and combined contributions of the disentanglement (factorization) and masking modules, we perform ablation experiments on the SDv1.4 training dataset. In this evaluation, overall performance (OP) refers to the average performance across all testing datasets, while generalization performance (GP) denotes the average performance on GAN-based datasets. The results for both \textit{ACC} and \textit{AP} are reported in Table~\ref{tab:module_ablation}.

When both modules are disabled, the framework reduces to the baseline UnivFD, which achieves 65.37\% (\textit{ACC}) and 75.31\% (\textit{AP}) in overall performance, and 60.42\% (\textit{ACC}) and 67.09\% (\textit{AP}) in generalization performance. Introducing the disentanglement module alone leads to significant improvements of +14.05\% (\textit{ACC}) and +14.22\% (\textit{AP}) in overall performance compared to the baseline, showing that separating causal and non-causal components helps the model capture task-relevant forensic cues. The masking module alone also brings improvements of +5.36\% (\textit{ACC}) and +6.82\% (\textit{AP}) in overall performance, demonstrating its ability to suppress spurious correlations. When the two modules are combined, the framework achieves the best results, with 89.64\% (\textit{ACC}) and 96.92\% (\textit{AP}) in overall performance, and 88.45\% (\textit{ACC}) and 95.25\% (\textit{AP}) in generalization performance. This corresponds to absolute gains of +24.27\% (\textit{ACC}) and +21.61\% (\textit{AP}) over the UnivFD baseline. These results confirm that disentanglement and masking are complementary, with disentanglement isolating stable causal features and masking further refining the feature space by removing style-specific noise.
\section{Conclusion}
In this paper, we introduce CausalCLIP, a causal representation learning framework aimed at improving the generalization of generated image detectors across diverse generative models. CausalCLIP disentangles representations through a Gumbel-Softmax-based masking mechanism guided by HSIC constraints to isolate causal forensic features from spurious ones. The adversarial masking strategy enables targeted suppression of non-causal components while preserving generalizable cues. Extensive experiments show our approach consistently outperforms existing methods, especially in cross-model settings. 
These findings highlight the importance of causal feature separation for generalization under distribution shifts. 
CausalCLIP provides a strong foundation for future image forensics research.

\section{Acknowledgments}
This work was supported in part by the National Natural Science Foundation of China under Grant 62376046, Grant U24B20182, and Grant 62561160098, and in part by the Natural Science Foundation of Chongqing under Grant CSTB2023NSCQ-MSX0341.


\bibliography{aaai2026}

\end{document}